\documentclass{article}

\usepackage{arxiv}

\usepackage[utf8]{inputenc} 
\usepackage[T1]{fontenc}    
\usepackage{hyperref}       
\usepackage{url}            
\usepackage{booktabs}       
\usepackage{amsfonts}       
\usepackage{amsmath}        
\usepackage{nicefrac}       
\usepackage{microtype}      
\usepackage{lipsum}		
\usepackage{graphicx}
\usepackage{subfigure}
\usepackage[sort, numbers]{natbib}
\usepackage{doi}

\title{Knowing the Distance: Understanding the Gap Between Synthetic and Real Data For Face Parsing}

\author{ {\hspace{1mm}Eli Friedman} \\
	\texttt{eli.friedman@datagen.tech} \\
	\And
	{\hspace{1mm}Assaf Lehr} \\
	\texttt{assaf.lehr@datagen.tech} \\
        \And
	{\hspace{1mm}Alexey Gruzdev} \\
	\texttt{alexey.gruzdev@datagen.tech} \\
        \And
	{\hspace{1mm}Vladimir Loginov} \\
	\texttt{vladimir.loginov@datagen.tech} \\
         \And
	{\hspace{1mm}Max Kogan} \\
	\texttt{max.kogan@datagen.tech} \\
        \And
	{\hspace{1mm}Moran Rubin} \\
	\texttt{moranrubin48@gmail.com} \\
        \And
	{\hspace{1mm}Orly Zvitia} \\
	\texttt{orly.zvitia@datagen.tech} \\
}
\date{}

\renewcommand{\headeright}{}
\renewcommand{\undertitle}{}
\renewcommand{\shorttitle}{}

\hypersetup{
pdftitle={Synthetic Data},
pdfsubject={Synthetic Data},
pdfauthor={Eli Friedman},
pdfkeywords={Synthetic Data, Datagen},
}

\begin{document}
\maketitle

\begin{abstract}
The use of synthetic data for training computer vision algorithms has become increasingly popular due to its cost-effectiveness, scalability, and ability to provide accurate multi-modality labels. Although recent studies have demonstrated impressive results when training networks solely on synthetic data, there remains a performance gap between synthetic and real data that is commonly attributed to lack of photorealism. The aim of this study is to investigate the gap in greater detail for the face parsing task. We differentiate between three types of gaps: distribution gap, label gap, and photorealism gap. Our findings show that the distribution gap is the largest contributor to the performance gap, accounting for over 50\% of the gap. By addressing this gap and accounting for the labels gap, we demonstrate that a model trained on synthetic data achieves comparable results to one trained on a similar amount of real data. This suggests that synthetic data is a viable alternative to real data, especially when real data is limited or difficult to obtain. Our study highlights the importance of content diversity in synthetic datasets and challenges the notion that the photorealism gap is the most critical factor affecting the performance of computer vision models trained on synthetic data.
\end{abstract}

\section{Introduction}

Two components are required to achieve successful results in computer vision tasks: an appropriate model and the right data. While significant efforts have been made in recent years to optimize models for solving complex computer vision tasks, there is also a growing focus on optimizing the data itself \cite{datacentric1, datacentric2}. Synthetic data provides a promising direction for data-centric optimization of computer vision models. Recent works that use synthetic data for computer vision tasks show impressive performance \cite{fake_it, digiface, nvidia}. Yet in order to further improve models trained on synthetic data, it is helpful to understand the potential gaps between synthetic and real data. We break down these differences into three types.  

\begin{figure}
    \centering
    \includegraphics[width=0.45\linewidth]{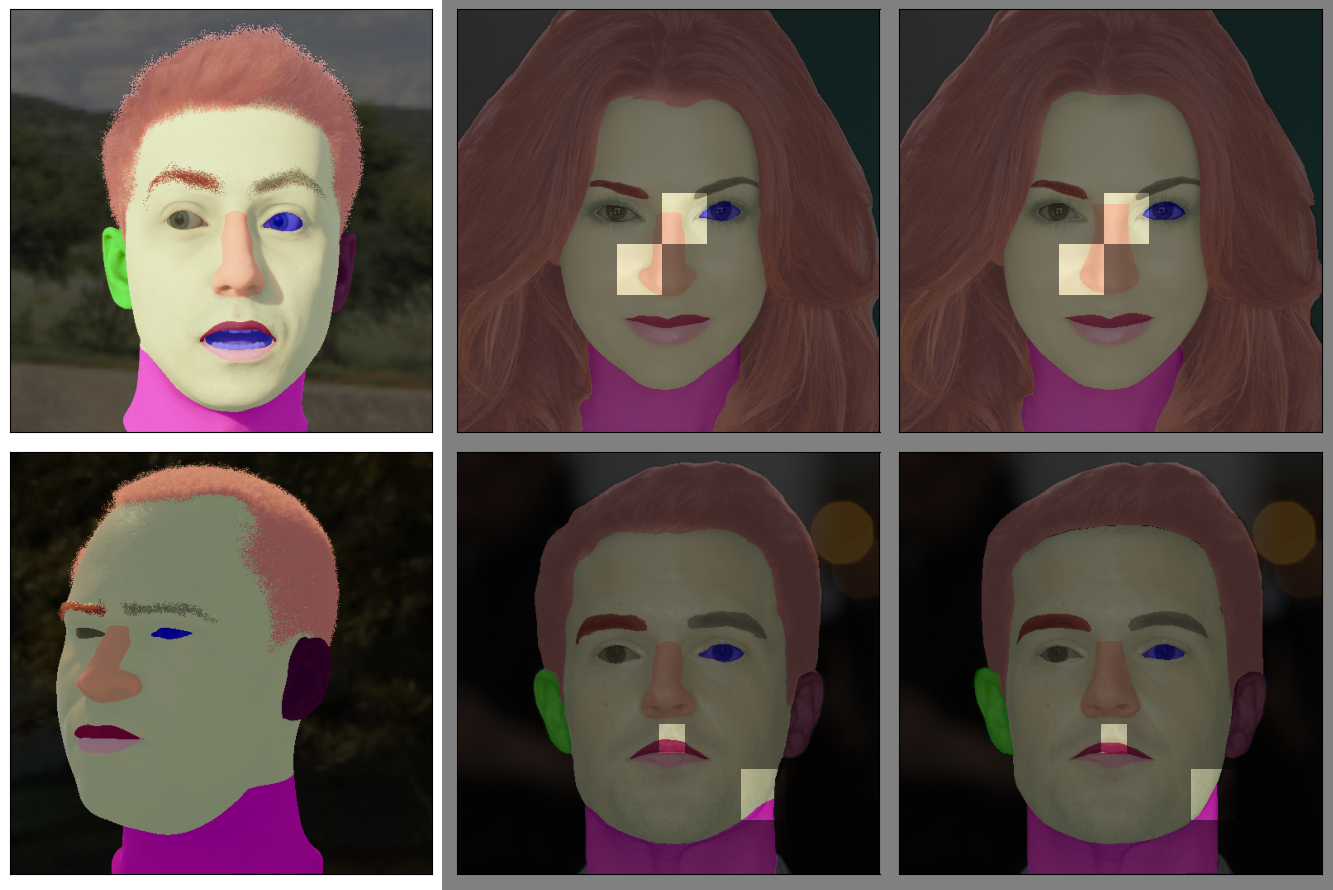}
     \caption{ \textbf{Label Gap} - A comparison of the synthetic labels and the CelebAMask labels. The left column shows the synthetic ground truth labels overlaid on synthetic images. The middle column shows accurate label predictions from our model which was trained on synthetic data and then applied to real images. The right column shows the ground truth labels. The highlighted regions show the differences in labeling conventions that are most noticeable---in the area of the nose, lips, and neck.
     }
\label{fig:label_gap_explained}
\end{figure}
\begin{figure}[t]
    \centering
        \subfigure{\includegraphics[width=0.2\linewidth]{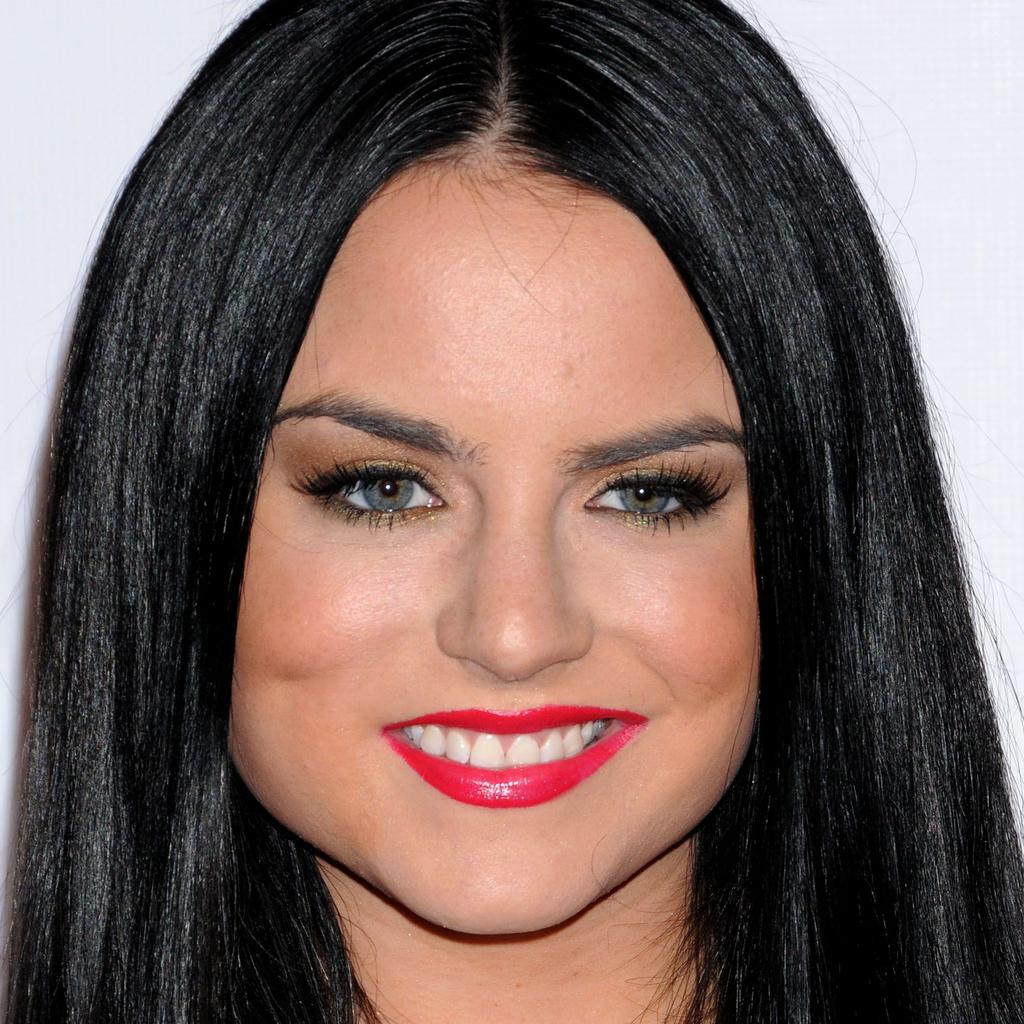}}
        \subfigure{\includegraphics[width=0.2\linewidth]{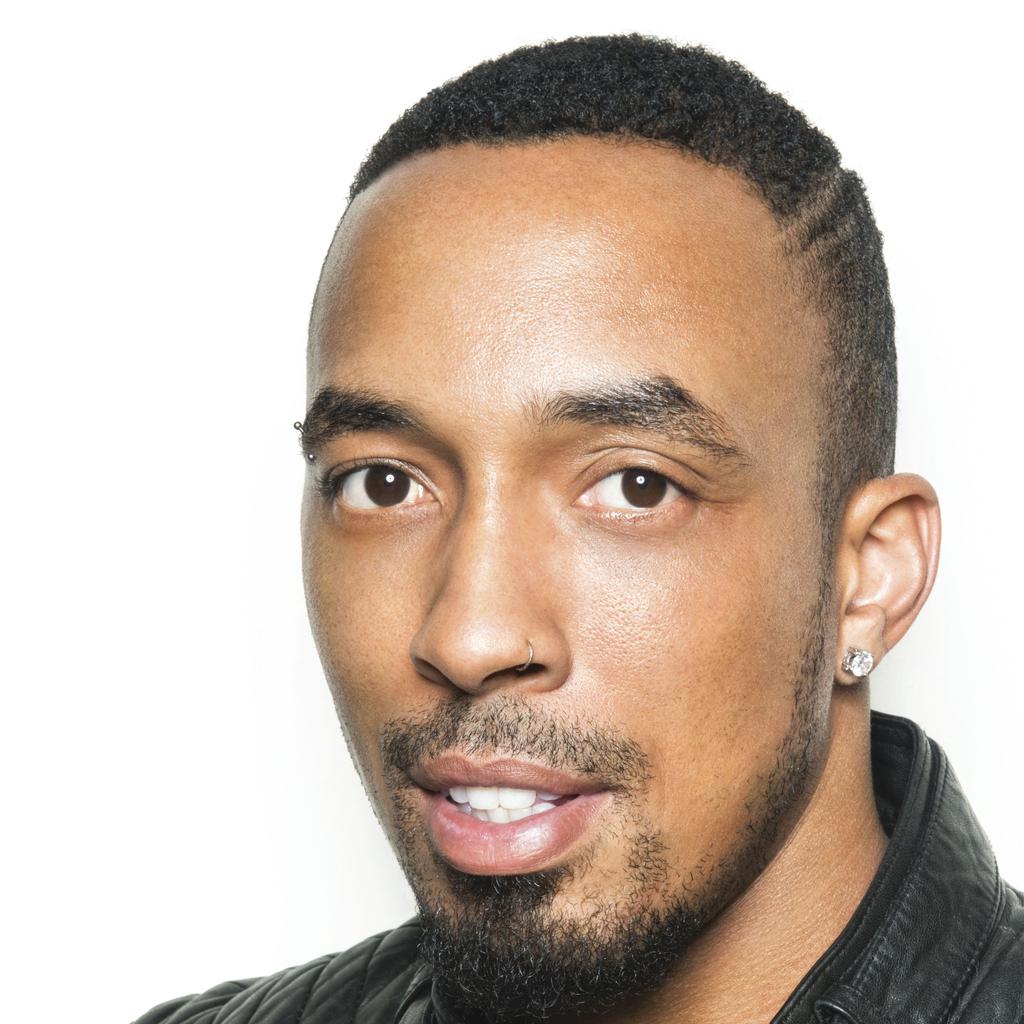}}
         \hfill \vrule \hfill
        \subfigure{\includegraphics[width=0.2\linewidth]{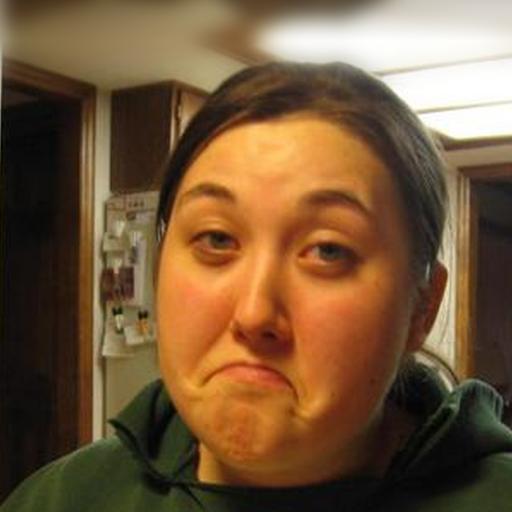}}
        \subfigure{\includegraphics[width=0.2\linewidth]{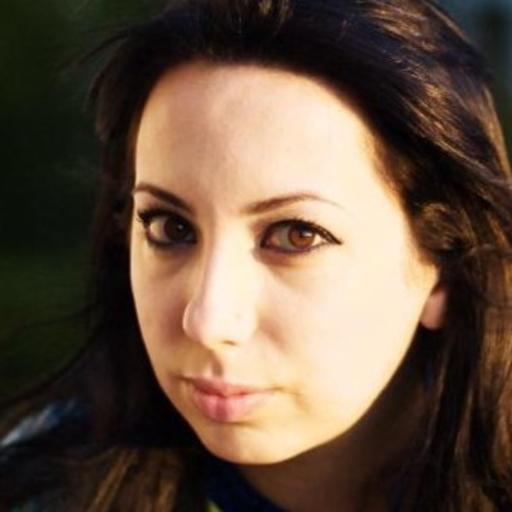}}

     \caption{\textbf{Visual Differences} - Consider the visual domain gap between the CelebAMask dataset examples (left) and the Helen dataset images \cite{helen} (right), which differ in lighting quality and color spectrum. This can occur between real datasets. When we discuss visual differences between synthetic and real datasets, we refer to it as the \textbf{photorealism gap}}.
     \label{fig:visual_gap_explained}
\end{figure}

\begin{itemize}
    \item \textbf{Distribution Gap} - This gap accounts for differences in the distribution of the content between two datasets. These differences can manifest in multiple ways. To name a few, there can be differences in object frequencies (one dataset may have a gender bias, age bias, or ethnicity bias), object scale (faces closer to the camera in one dataset than the other), and the absence of certain elements in the training data that may be present in the testing scenario (e.g., hats, earrings and other accessories). Using datasets with different distributions for training and testing can lead to a variance deficiency in the training distribution relative to the test distribution, which can negatively impact the performance of the model. While this problem may also occur whenever using two \textbf{real} datasets, if one is using synthetic data, then the distribution gap is addressable, either by adapting the generation parameters to align more closely with the real data, or by creating additional 3D assets that the dataset lacks.
    \item \textbf{Label gap} - This gap arises due to inconsistencies in labeling conventions for the same semantics. For instance, two datasets might have different conventions about where the nose ends and the skin begins. The label gap may similarly occur between two real datasets, if they are labeled based on different conventions. This leads to an evaluation challenge as a model trained on a dataset with labeling-instructions-set A may show significant error rates when tested on data tagged with instructions-set B, even if it has performed perfectly on a test set with the training dataset's conventions. See Figure \ref{fig:label_gap_explained} for an example of the label gap between the real and our synthetic labels.
    \item \textbf{Photorealism gap} - This accounts for any image level visual differences between real and synthetic data, such as image noise, color variations, texture differences, or other discrepancies. The photorealism gap is a specific type of visual domain gap, which can also occur between two real datasets due to factors like differences in camera sensors or lighting conditions. See Figure \ref{fig:visual_gap_explained} for an example of a visual domain gap between real datasets. We refer to the visual domain gap between synthetic and real data as the photorealism gap, which occurs when a synthetic image lacks the realism of a photograph.
\end{itemize}

 This paper investigates the gaps between synthetic and real data in the context of face parsing, which involves segmenting an image into distinct regions corresponding to different facial areas. Specifically, we use synthetic data to train a model for this task and compare its performance to models trained on real-world data. 
 
 Face parsing is a challenging task, as faces can vary in appearance, pose, coloring, and images can vary significantly in lighting, occlusions and accessories. Collecting a large dataset is difficult due to privacy concerns and labeling efforts. Additionally, the initial collected dataset may not cover all necessary test scenarios, and we may need multiple rounds of data collection and annotation to achieve optimal results. This iterative process of improving a dataset is called \textbf{data-centric iterations}. 
 However, when using synthetic data, we can avoid the need for iterative data collection by generating multiple controlled datasets that quickly converge to the required data distribution. 
 With 3D simulated data, controlling distribution gaps such as object frequency and variety, occlusions, and camera viewpoint can be relatively easy with pre-existing 3D assets. It is also possible to create new assets on demand to reduce content gaps. The photorealism gap due to texture differences or camera noise can also be mitigated although it might be more challenging to accurately identify and simulate the test domain.
 We train a model using synthetic data and test it on the challenging CelebA-Mask dataset. 
 We show that synthetically simulated face data offers a potential solution to the shortage of labeled data for face parsing tasks.

The paper is structured as follows: Section \ref{section:related_work} reviews prior research in the field. Section \ref{section:method} outlines our method and training, while Section \ref{section:experiments} presents our results. In Section \ref{section:discussion}, we interpret these results, and in Section \ref{section:future_work}, we suggest directions for future work.

\subsection{Contributions}
The contributions of this paper are as follows: 
\begin{itemize}
    \item We provide a framework for understanding the performance gap between real and synthetic data.
    \item We provide evidence that the distribution gap, rather than the photorealism gap, makes up the largest portion of the performance gap.
    \item We demonstrate that a model trained purely on synthetic data can achieve comparable results to real data for face parsing tasks
    \item We demonstrate the advantage of using the accurate synthetic labels over human-annotated labels for dense segments like hair
\end{itemize}

\section{Related Work}
\label{section:related_work}

In the section we cover prior work related to our current research: real and synthetic face parsing datasets, label adaptation and domain adaptation.  
\subsection{Face Parsing Models}
Face parsing is the process of segmenting a person's face into various sections. There are several model architectures and training methods available for face parsing, including pretraining on large image-language datasets \cite{face_parsing1}, using transformers \cite{face_parsing1, face_parsing2}, and graph methods \cite{face_parsing3}. However, our focus is on data design rather than model architecture, hence, we follow \cite{fake_it} and use a simple UNet \cite{unet}, which works well in practice.

\subsection{Face Parsing Datasets}
\textbf{Real Datasets}
There are several publicly available datasets for face parsing, including Helen \cite{helen}, which contains 2,330 images, \cite{LaPa}, with 22,176 images, and CelebAMask \cite{celeba}, which contains 30,000 high quality images collected from the larger CelebA dataset \cite{celeba_full}. The F$_1$ score is the most standard metric for evaluating face parsing models since it takes into account the variation in sizes between different labels and weights each class equally. The F$_1$ score is calculated as: 
\begin{equation*}
 F_1 = \frac{1}{\# \textrm{labels}}\sum\limits_{label} 2 \frac{\textrm{precision}_{label} * \textrm{recall}_{label}}{\textrm{precision}_{label} + \textrm{recall}_{label}}    
\end{equation*}

\textbf{Synthetic Face Datasets}
The task of collecting face data for various face-related tasks, including face recognition, face parsing, and face landmark detection, poses a significant challenge due to potential privacy concerns. To overcome this challenge, researchers have explored the use of synthetic face data generated through 2D generative models \cite{2D_facegen}, or simulated 3D techniques \cite{fake_it}. \citet{2D_facegen} use the StyleGAN network \cite{stylegan} to generate privacy friendly synthetic dataset for face parsing. \citet{fake_it} use a 3DMM and a graphics rendering engine to generate a large dataset of 100,000 simulated synthetic face images for landmark detection and face parsing. They demonstrate that after label adaptation, they achieve competitive results to real data on the LaPa \cite{LaPa} and Helen* datasets \cite{helen, face_parsing2}.  These techniques have proven effective in enhancing the performance of deep learning models, as they provide a diverse and easily controllable dataset without privacy concerns.

\subsection{Label Adaptation}
Consider a model learning on one dataset, but being evaluated on a test set where the annotators were given very different instructions for labelling. Even if the model performs well on images labelled use the training data's conventions, the score on the test data may not accurately reflect the success of the model since the labels on the datasets differ.
Label adaptation \cite{fake_it} is a technique used to compare and evaluate models trained on different datasets that may have different labeling conventions for the same semantic classes. Label adaptation adjusts the predicted labels from a model to align with the labeling conventions of another dataset. In our context, we train a model on synthetic data and then use label adaptation to evaluate how well our model works on a real-world dataset with different labeling conventions.

The label adaptation process involves two steps. First, the trained model is run in inference mode on the real-world dataset to obtain labels using the synthetic dataset's convention. Second, another model is trained to translate the synthetic predictions to the labels used in the real-world dataset.

\subsection{Domain Adaptation}
In machine learning, domain adaptation aims to enhance the performance of a model that was trained on one domain when it is used on a different domain. Various domain adaptation frameworks handle dataset differences by focusing on variations in image distributions, known as covariate shift \cite{DA_Survey1, DA_Survey2}. Other domain adaptation works focus on solving the photorealism gap and develop techniques that adapt image appearances to more resemble real images \cite{visual_domain1, visual_domain2, visual_domain3}. However, they don't distinguish between different types of distribution divergences.

Perhaps the simplest technique for adapting a model trained on one domain to another is simple fine-tuning. This process involves taking a pre-trained model that has been trained on synthetic data and then fine-tuning it on a small amount of real data. The idea is to adjust the weights of the model to better fit the distribution of the real data.The advantage of fine-tuning is that it requires very little additional real data and can be done relatively quickly \cite{handsup}. Fine-tuning with real data can potentially overcome all the aforementioned gaps. However, the amount of data needed may vary depending on the severity of the gaps, and in some cases it might be difficult to obtain enough data.

When designing synthetic datasets for face parsing, it is crucial to consider the distribution of the data. For example, each image contains a face and a decision needs to be made about what ethnicity, gender, and age the identity needs to have, along with their hairstyle, hair color, and eye color. The face needs to be positioned, and the parameters for the camera need to be chosen. There are many free variables, and it is not obvious which ones to choose. 
\citet{nvidia} establish the importance of matching the distribution of content in synthetic datasets to that of real datasets, and introduce a novel technique for achieving this. They are able to effectively match the distributions of the two datasets by backpropagating the Maximum Mean Discrepancy loss through the rendering engine to the probabilistic graph that generates the scenes. Their work showed that aligning content distributions can make synthetic data useful for real-world tasks. We build upon this idea and present further evidence in support of it.
While it is true that an automatic method such as \citet{nvidia} for setting the parameters could potentially improve results, implementing such a method can be complex and time-consuming. Therefore, we opted to manually and iteratively choose the parameters for our study. This works reasonably well in achieving our objectives.

\section{Method}
\label{section:method}

\subsection{Datasets}
We measured our results on the CelebAMask dataset \cite{celeba}, a challenging face parsing dataset that contains 30,000 annotated images: 24,183 for training, 2,993 for validation and 2,824 for testing. We chose this dataset as it contains the largest number of images and the faces are pre-aligned in the image. The CelebAMask contains 19 categories, but we measure the F$_1$ score averaged over all the categories excluding the earrings, clothes, and necklace, for a total of 16 categories. We exclude these categories since the CelebAMask dataset overlays them on top of the other facial labels, and we focused on measuring our ability to segment facial areas only.
We resized the CelebAMask images using bilinear interpolation from 1024x1024 to 512x512, which is the size at which it was originally labeled \cite{celeba}.
 
To investigate the differences between synthetic and real data on a face parsing task, we generated data using Datagen's face generation platform \cite{handsup} and created a synthetic training dataset of widely varied face images. We generate each image at 512x512 resolution. The face-generation platform uses a physically-based rendering engine that renders 2D images from 3D models. It also produces additional 3D data, such as key points, normal maps, depth maps and segmentation maps. \footnote{For more information about our platform 
see \url{https://datagen.tech}}

\textbf{Generator Parameters:}
We began by generating a dataset of 22,488 images. 
We uniformly sampled age, ethnicity, and gender from a database of tens of thousands of distinct identities,
while using the default hair and eye parameters for each identity. It's worth noting that this uniform sampling approach could potentially lead to less accurate results in case of imbalanced representation of certain attributes. For instance, our synthetic dataset contains 50\% females, whereas in CelebAMask, only 37\% of the images contain females as measured by the CelebA annotations \cite{celeba_full}.
As a result, it's possible that using CelebAMask as a test dataset may not accurately reflect a potential real-world use case. We sample the camera position, location, and field of view so that the faces occupy 200 - 300 pixels in the image and the angle of the face is distributed according to a truncated 
Gaussian distribution centered on $0^{\circ}$ and limited to $\pm90^{\circ}$. This means most of the faces are front facing, as in the CelebAMask dataset, but with some extreme poses up to $90^{\circ}$. The CelebAMask dataset lacks a beard label, and instead estimates the jaw line to mark the skin label. To maintain consistency with this labeling convention, we render each image with a beard twice---once with a beard and once without---and use the image with the beard and the segmentation without the beard. This approach enables us to incorporate beards in our images while only segmenting the face underneath. See the first column in Table \ref{tab:dataset_breakdown} for the distribution of the initial dataset. We chose 22.7\% of the faces to have no expression, and the rest we gave a randomly chosen expression from the following expressions: fear, anger, contempt, happiness, disgust, surprise, and sadness.

\begin{figure}
	\centering
        \subfigure{\includegraphics[width=0.17\linewidth]{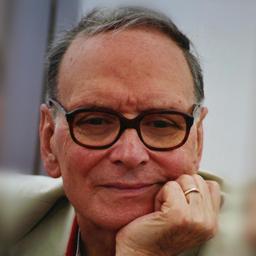}}
        \subfigure{\includegraphics[width=0.17\linewidth]{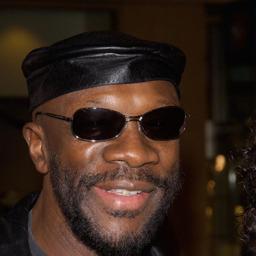}}
        \subfigure{\includegraphics[width=0.17\linewidth]{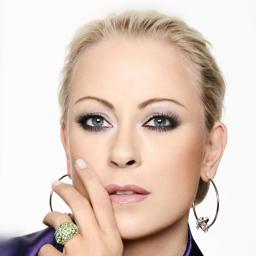}}
        \subfigure{\includegraphics[width=0.17\linewidth]{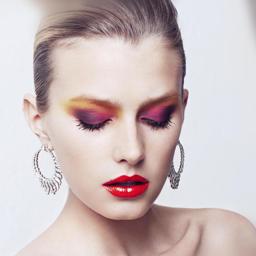}}
        \subfigure{\includegraphics[width=0.17\linewidth]{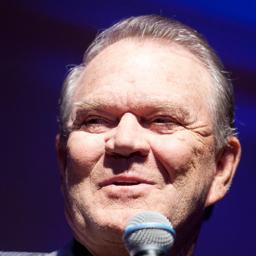}}
	\caption{\textbf{Real samples} - Some examples from the CelebA dataset that include hats, makeup, earrings, and occlusions. These are more challenging cases for the model.}
	\label{fig:content_examples}
\end{figure}

\begin{figure}
	\centering
        \subfigure{\includegraphics[width=0.17\linewidth]{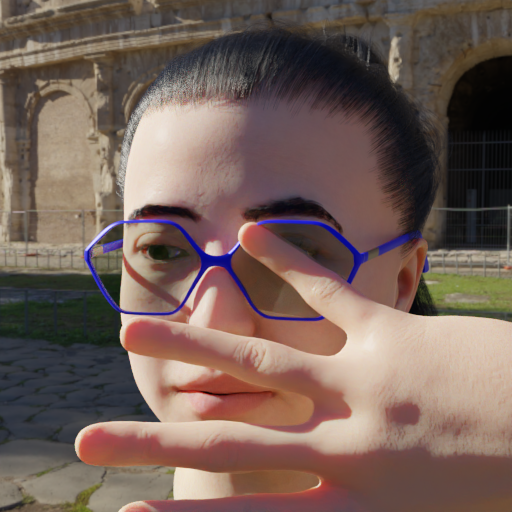}}
        \subfigure{\includegraphics[width=0.17\linewidth]{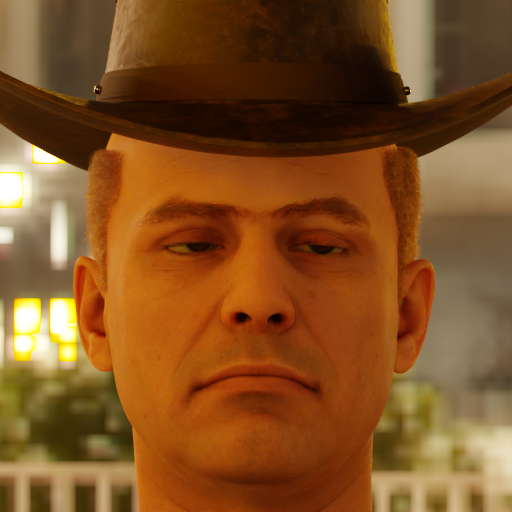}}
        \subfigure{\includegraphics[width=0.17\linewidth]{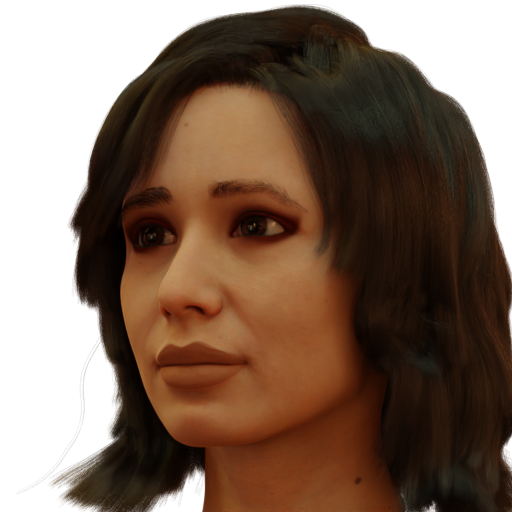}}
        \subfigure{\includegraphics[width=0.17\linewidth]{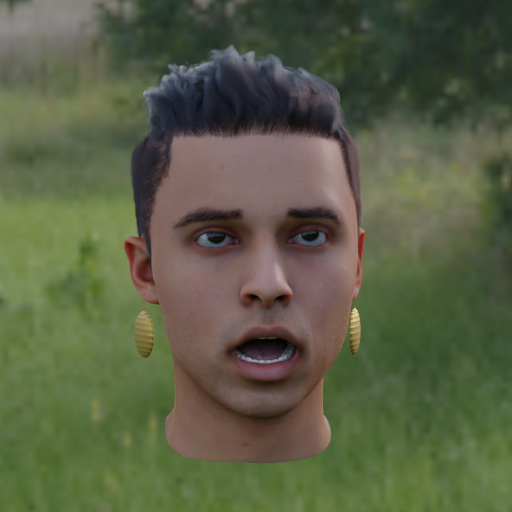}}
        \subfigure{\includegraphics[width=0.17\linewidth]{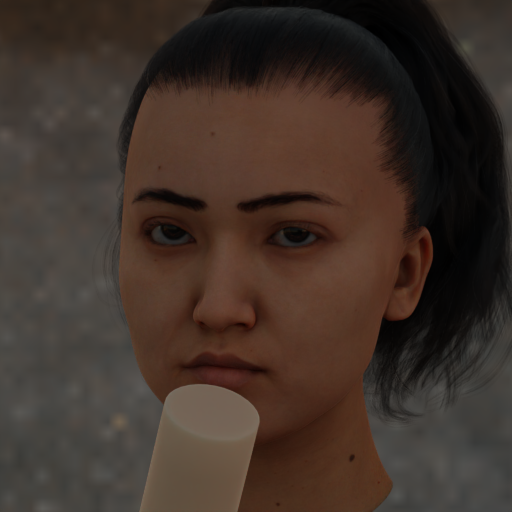}}
	\caption{\textbf{Synthetic samples} - Examples of images added to the synthetic dataset to better align the distribution to that of the CelebAMask dataset. Accessories include hats, makeup, earrings, and occlusions caused by hands and objects.
 }
	\label{fig:synthetic_content}
\end{figure}

\subsection{Iterative Approach}
Following our initial training, we iteratively improved our model by performing these steps: 
\begin{enumerate}
    \item \textit{Generate a synthetic dataset.} We chose parameters to try and maximize variance while attempting to stay true to the distribution of the real data. 
    \item \textit{Train UNet model on synthetic dataset.} It is critical to aggressively apply augmentations to the synthetic data.
    \item \textit{Run inference on real data.} 
    \item \textit{Analyze systematic errors.} We look for patterns in the error cases to understand what data is missing that is causing the model to make mistakes. Once we know what's missing, we can generate new data to make up for the lack in the original dataset by repeating the procedure.
\end{enumerate}

Throughout our error analysis steps 
we found multiple errors which we fixed by generating adequate datasets. For example, we noticed the absence of hats, eyeliner, and earrings in our initial dataset. In addition, real-world images contained occlusions not present in our data at first as people often put their hands over their faces or hold objects in front of them. See Figure \ref{fig:content_examples} for some examples. The promise of rendered synthetic data is that closing these types of gaps and fixing edge cases is easier than collecting and annotating new images. This of course depends on the availability of relevant 3D assets to close the distribution gap. In our case, we generated new images containing the missing accessories and makeup. We also added images with occlusions by adding 3D objects in the scene randomly placed in front of the face, or moving the person's hands to occasionally cover the face. See Figure \ref{fig:synthetic_content} for some images of the variance added to the dataset.

\begin{table}
	\caption{Synthetic dataset breakdown between the initial synthetic dataset and the dataset after adding additional variance to align the distributions better. Each column shows the fraction of the synthetic dataset that contains each type of content.}
	\centering
	\begin{tabular}{lll}
		\toprule
		\cmidrule(r){1-3}
		Variance Type & \% of Initial Dataset & \% of Synthetic Dataset + Variance \\
		\midrule
    	Background Images & 100,000 & 100,000 \\
		\midrule
            Daytime & 72.0\% & 72.0\% \\
            Evening & 14.2\% & 14.2\% \\
            Night & 13.8\% & 13.8\% \\
    	Earrings & 0.0\% & 8.8\% \\
            Beard & 9.0\% & 9.0\% \\
    	Makeup & 0.0\% & 8.4\% \\
            Hat & 0.0\% & 16.02\% \\
            Glasses & 18.6\% & 18.6\% \\
            Extreme Pose & 13.9\% & 13.9\% \\
    	Occlusions & 0.0\% & 17.4\% \\
    	  Closed Eyes & 4.8\% & 4.8\% \\
    	  Mouth Open & 4.7\% & 4.7\% \\
		\bottomrule
		Total & 22,488 Images & 22,488 Images \\
	\end{tabular}
	\label{tab:dataset_breakdown}
\end{table}

To make it easier to compare the performance of the new dataset with additional variance to the previous version, we replaced old images in the dataset with the new ones. This way, we kept the same number of images in the dataset. Table \ref{tab:dataset_breakdown} shows the data distribution of the two datasets.

\subsection{Training}

\textbf{Model:} Since our focus is on optimizing the training data, rather than on optimizing the model, we follow \citet{fake_it} and use a simple UNet \cite{unet} with a Resnet-18 backbone \cite{resnet, resnet_implementation}. The input to the network is a 512x512 RGB image and the output is a 16 channel segmentation map for each of the face regions plus the hat category. We trained all models until convergence using the Adam optimizer with a fixed learning rate of $10^{-4}$, $\beta_1 = 0.9$ and $\beta_2 = 0.99$ and a batch size of 32.

\textbf{Augmentations:}
The process of rendering 3D images to create synthetic data assumes a perfect camera model, resulting in noise-free images. However, in the real world, cameras capture images with varying degrees of noise and lighting conditions, making augmentations critical for achieving optimal results when training with synthetic data. To this end, we employ a diverse range of augmentations, elaborated in Appendix \ref{appendix:augmentations}, to improve the generalization capabilities of the model. When training our reference model on the CelebAMask dataset, we applied the same augmentations, and found that they were also helpful for improving the results on the real data.

To further enhance the robustness of our approach, we use Stable Diffusion \cite{StableD} to generate 100,000 background images. During training, we use the rendered alpha mask of the face to composite these images with the face images. By doing so, the model can learn to ignore various background objects, resulting in improved performance and greater robustness. Since the alpha masks are not available for the real data, we were only able to apply the randomized backgrounds to the synthetic data.

\textbf{Label Adaptation:}
\begin{figure}
    \centering
    \includegraphics[width=0.6\linewidth]{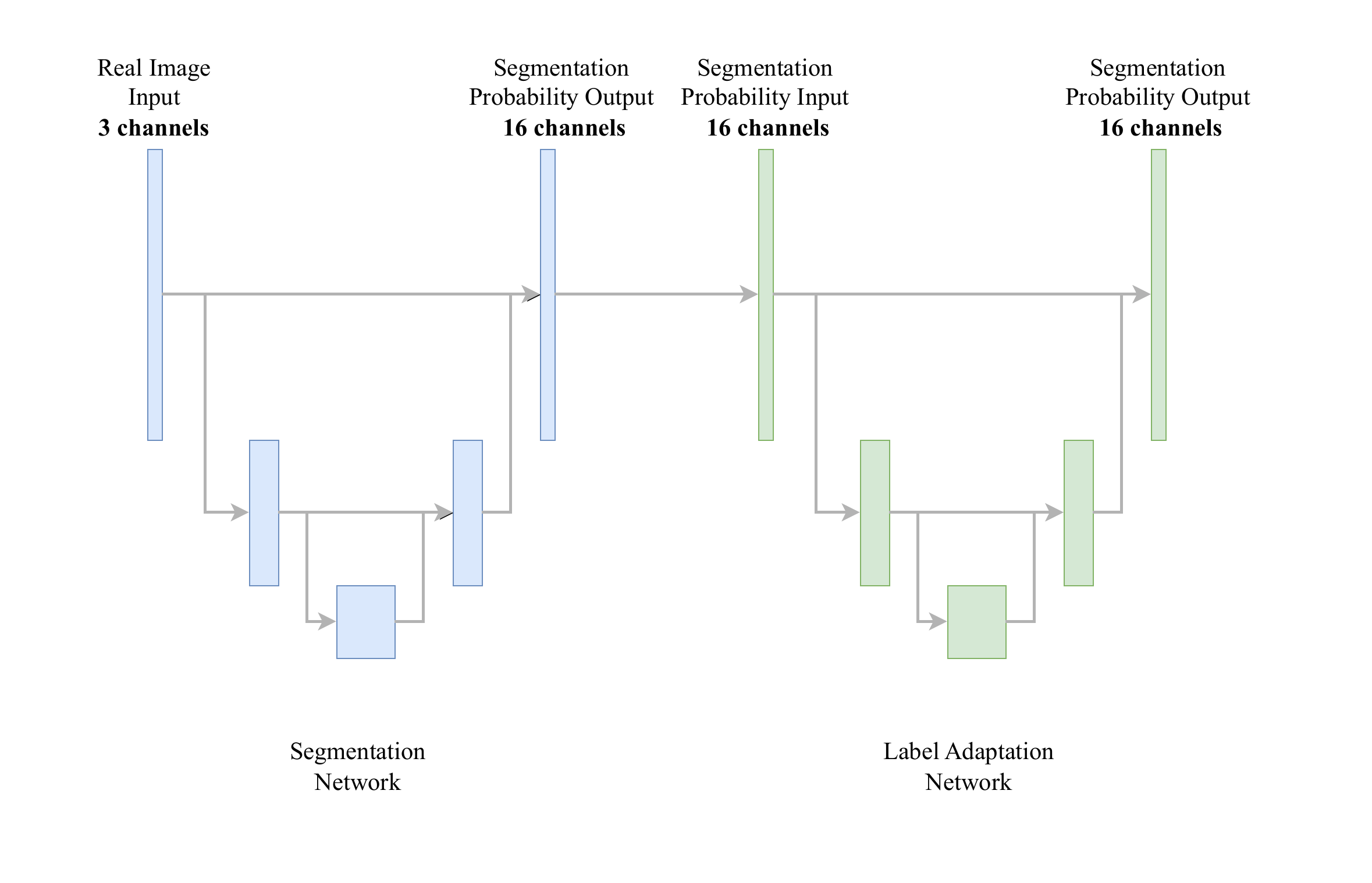}
    \caption{\textbf{Label Adaptation:} A segmentation model that was trained on synthetic data (blue) is frozen and applied on real RGB images. This network outputs a 16 channel image containing segmentation probabilities in the convention of the synthetic data, which are input to the label adaptation network (green). The label adaptation network is trained to correct these and outputs segmentation labels in the convention of the real data.
    }
\label{fig:label_adaptation_diagram}
\end{figure}
Our label adaptation model shares the same UNet architecture as the segmentation model, except the input layer takes the probabilities from the segmentation model as input rather than an RGB image. In order to speed up training, we initialize the weights of the label adaptation model using the frozen segmentation model's weights, except for the input layer which is initialized randomly. By using the label adaptation model, we can compare the performance of our synthetically trained model to a model solely trained on real data. We conduct experiments to determine the influence of dataset size on training the label adaptation model and how much real data is required for optimal results. See the label adaptation architecture in Figure \ref{fig:label_adaptation_diagram}. The label adaptation network was trained using the same optimization parameters as the segmentation model, except that the batch size was reduced to 16. For inference, we chose the model checkpoint that maximized the F$_1$ score on the validation set.

\textbf{Fine-tuning:}
To better understand the impact of real data on correcting the photorealism gap, we conduct experiments where we fine-tune the model using different amounts of real data. By doing so, we aim to gain insights into how much real data is needed to overcome any remaining variance or photorealism gap. We began with a model trained using synthetic data, and then train for another few epochs on a dataset of real images from the CelebAMask training set. We use the same hyperparameters as the initial training (described above) and chose the model checkpoint that maximized the F$_1$ score on the validation set.

\section{Results}
\label{section:experiments}
In order to evaluate the relative influence of the different gaps we run the following experiments. 

\subsection{Understanding the Distribution Gap}
We compare the results of our initial dataset to the one after our iterative improvements. See Table \ref{tab:dataset_breakdown} for a comparison of the two datasets.
Table \ref{tab:dataset_results} shows that the additional variance added to the dataset increases the F$_1$ score by 5.6 percentage points to 86.3\%. It should be noted that a portion of this improvement is attributed to the hat category, as our initial dataset lacked any instances of hats, and was evaluated using the hat category. Nevertheless, it is still significant that the addition of even a limited amount of new content leads to a notable increase in the score. Also worth noting that this improvement is observed despite the known label gaps.

\begin{table}
	\centering
	\begin{tabular}{lll}
		\toprule
		\cmidrule(r){1-2}
		Dataset & F$_1$ score (\%) & Improvement (\%)\\
		\midrule
		  Initial Synthetic dataset & 80.7 & \\
		  Synthetic With Variance & 86.3 & +5.6\\
		  Synthetic With Variance + Label Adaptation & 90.1 & +9.4\\
            \midrule
            CelebAMask & 91.2 & \\
		\bottomrule
	\end{tabular}
	\caption{Results of different training datasets. We compare training with our initial synthetic dataset, our synthetic dataset with additional variance, and with label adaptation. We compare to the reference CelebAMask dataset.}
	\label{tab:dataset_results}
\end{table}

\subsection{Understanding the Label Gap}
\begin{figure}
	\centering
	\includegraphics[width=0.75\textwidth]{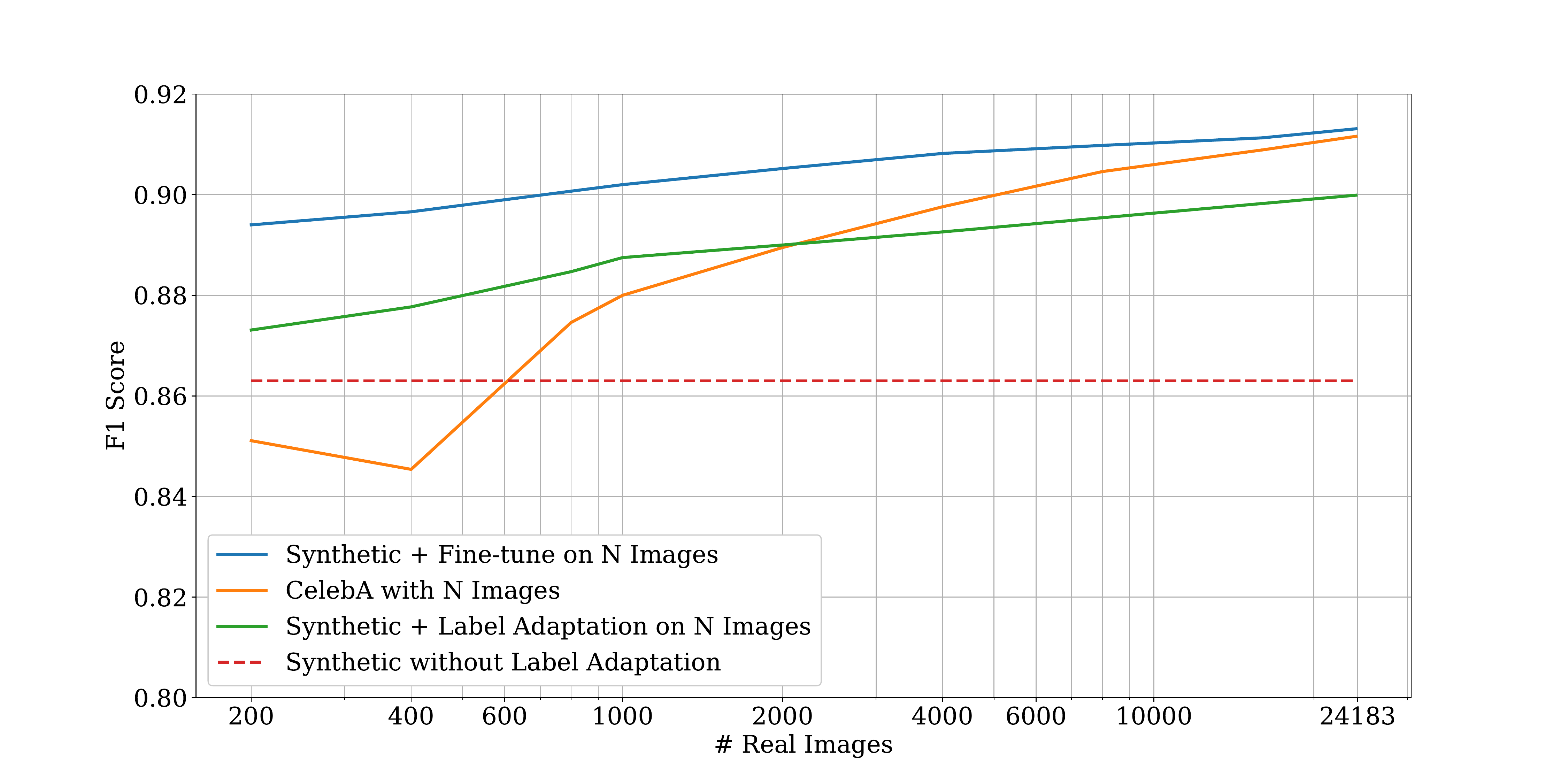}
	\caption{We compare the F$_1$ scores of different models: a model trained on synthetic data and then fine-tuned on varying amounts of real images (blue), a model trained on synthetic data with a label adaptation model trained on varying amounts of real images (green), a model trained solely on varying amounts of real images (orange), a model trained only on synthetic data without label adaptation (dashed red). The same information is presented in table format in Appendix \ref{appendix:result_table} }
	\label{fig:results_comparison}
\end{figure}
To ensure a fair comparison between the model trained on synthetic data and one trained on real data, we also apply label adaptation. Table \ref{tab:dataset_results} shows the results of training with the additional variance and then adapting the labels using the full CelebAMask dataset. An additional 3.8 percentage point gain is achieved when accurately accounting for the difference in label conventions.
We also experiment with varying the amount of real images used in the label adaptation process.
Figure \ref{fig:results_comparison} demonstrates that label adaptation using even small amounts of real data can improve the training results on pure synthetic data. When 200 images are used, the F$_1$ score increases from 86.3\% to 87.3\%.
What's particularly interesting about the graph is what's shown on the right-hand side. By using the entire CelebAMask training set to refine the labels, we ensure that they are fully aligned with their real-world counterparts, allowing us to accurately measure how well the model trained on synthetic data performs on real data. The result is that synthetic data performs on a comparable level to the real data, with only a 1 percentage point difference. This finding underscores the potential of synthetic data as a reliable and effective alternative to real-world data in applications where the real and synthetic labels are well aligned. It should be noted that this improvement in F$_1$ score may not be solely due to the adaptation of labels, as the label adaptation process also corrects some errors in the model's predictions.

\subsection{Fine-tuning with Real Data}
\begin{table}
	\centering
	\begin{tabular}{llll}
		\toprule
		Experiment & CelebAMask Dataset & F$_1$ score (\%) \\
		\midrule
		  Train from scratch & 200 Minimal-variance Images & 75.9 \\
		  Train from scratch & 200 Randomly Chosen Images & 85.1 \\
		\midrule
		  Train on Synthetic Data \small{(no label adaptation)} & 0 Images & 86.3 \\
            \midrule
		  Fine-tune synthetic model & 200 Minimal-variance Images &  87.8 \\
		  Fine-tune synthetic model & 200 Randomly Chosen Images & 89.4 \\
		\bottomrule
	\end{tabular}
	\caption{We compare two real datasets: one without any additional variance, and one randomly selected. We train and fine-tune a model trained on synthetic data using these datasets. The randomly chosen dataset shows improved results both for training and fine-tuning, thus showing the importance of additional variance. For reference, we show the results of the synthetic dataset that was used for fine-tuning.}
	\label{tab:small_data_results}
\end{table}
In order to overcome the photorealism gap, we fine-tune our synthetic network with varying amount of real samples. As illustrated in Figure \ref{fig:results_comparison}, fine-tuning on real data consistently outperforms label adaptation. As previously mentioned, exposing the network to real data allows it to overcome all three types of gaps. The question that arises is then:  "what accounts for this difference?" It cannot be attributed to the content of the data, since in our experiment both fine-tuning and label adaptation were trained on the same datasets. Instead, we hypothesize that the network's direct exposure to RGB images during fine-tuning allows it to adapt to any visual discrepancies between the synthetic and real data, such as variations in texture, image noise, and lighting. In contrast, the label adaptation process only exposes the network to label probabilities, which may not provide sufficient information for the network to fully adjust to such visual differences. This highlights the relatively minor impact the photorelism gap has on the results, as it only accounts for, on average, a 1.6 percentage point increase above the label adaptation.

Another interesting observation is that fine-tuning a synthetic model with real data (blue curve) consistently outperform training a model from scratch using only real data of the same amount used for fine-tuning (orange curve). This emphasizes the value of synthetic data even when real data is available as it enables using significantly smaller amounts of real data for training.

In order to investigate the role of variance vs. photorealism in the performance gap, we conducted an additional experiment. We select two small real image datasets, each containing 200 images from the CelebAMask training set. One dataset consisted of 200 randomly sampled images, while the other contained 200 images that excluded all additional variance such as hats, glasses, earrings. The purpose of the second dataset is to expose the network to photorealistic images, but without exposing it to any variance. We fine-tuned our final synthetic network (after iterative improvements) using these two datasets and also used them to train two models from scratch. Table \ref{tab:small_data_results} presents the results. The fine-tuning was performed on a network trained on synthetic data that acheived an F$_1$ score of 86.3\%. Fine-tuning this network on the content-limited dataset increased the F$_1$ score 87.8\%, whereas fine-tuning it on the randomly chosen dataset increases the score increased to 89.4\%. These results are another indication that closing the content gap is more important than the photorealism gap and the label gap combined. The content-limited dataset increases the score by 1.4 percentage point  s and contains photorealistic images labelled using the real-data conventions, but does not contain any additional variance. The randomly chosen dataset, on the other hand, increases the score by 3.1 percentage points, and it contains high variance in the content, in addition to photorealistic data labelled using the real data conventions. This suggests that closing the distribution gap leads to as big an improvement as the closing the photorealism gap and the label gap combined.

The significant difference between training from scratch on the minimal variance dataset and fine-tuning is partly due to the extra variability in the synthetic dataset, which includes items like hats that are not present in the minimal variance dataset.

\subsection{Comparison to Other Synthetic Datasets}
We also compare our synthetic dataset to the Face Synthetics Dataset of \citet{fake_it}. Their dataset contains 100,000 synthetically generated faces and includes a wide variety of assets including clothes, hats, glasses, and masks. While we have access to their dataset, we did not have access to their data generator. This posed a challenge, since we could not correct the beard or mask segmentation as we did with our data. We trained using the same augmentations as we used for our dataset (see Appendix \ref{appendix:augmentations}) and the full Face Synthetics Dataset and achieved an F$_1$ Score 83\%. We also trained using using a subset of the full dataset that contained images without a beard or mask and achieved an F$_1$ Score of 83.7\%. 

\subsection{Labels Accuracy on Dense Hair Segments} 
\begin{figure}
	\centering
        \subfigure{\includegraphics[width=0.7\linewidth]{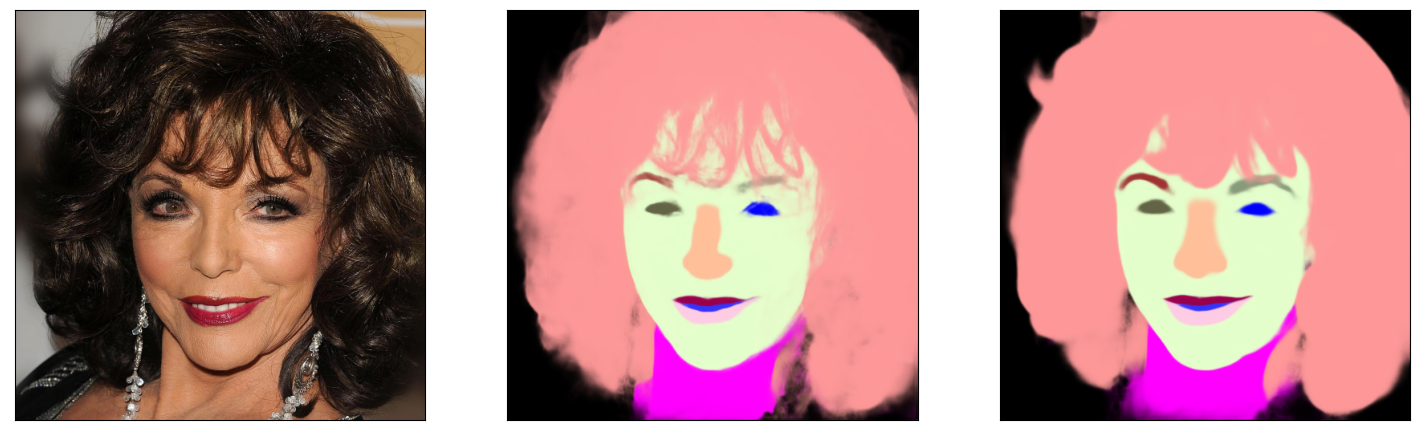}}
        \hspace{0.5cm}
        \subfigure{\includegraphics[width=0.7\linewidth]{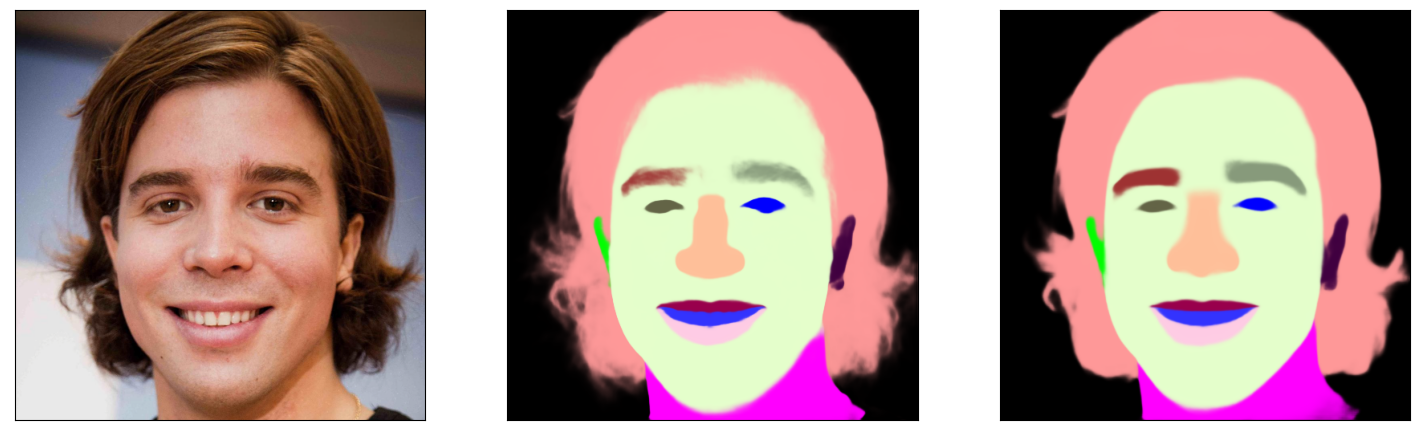}}
	\caption{The left column shows the RGB image, the center column visualizes the probability distribution of the network output trained on synthetic data, and the right column shows the probability outputs of the network trained on real data. It is worth noting that in areas where hair is present, the probabilities can be interpreted as the proportion of the pixel that is occupied by hair, thus providing subpixel accuracy for hair detection. }
	\label{fig:synthetic_hair}
\end{figure}
Synthetic labels are precisely aligned with the pixels in the corresponding images, resulting in a level of accuracy that may surpass what a human can achieve when labeling real data. This is especially important in finely detailed areas, such as hair. When using synthetic data, the hair is labelled even with subpixel accuracy---if, for example, 30\% of a pixel contains a hair, then it will be labeled as a hair with probability 30\%.  When a model is trained over a full dataset of these statistical labels, it ends up learning to correctly predict the probability that a pixel contains hair. In Figure \ref{fig:synthetic_hair}, we plot the output probabilities of the network, and blend the colors together weighted by their respective probabilities. The rightmost column shows the network output from a model trained on real data. We can observe that due to the human labelling in the right column, the network outputs are imprecise and cover larger regions than the actual hair. In contrast, the middle column displays the outputs of the synthetically-trained network, and the outputs align more closely with the strands of the person's hair. Not every individual strand is captured, however finer detail is achieved than what the human labels capture.
These precise labels may be beneficial when accurate segmentation of hair is desired (e.g., background separation or hair-dye try on).

\section{Discussion} 
\label{section:discussion}
Our study shows that synthetic data performs almost as well as real data, with a difference of only 1 percentage point once differences in label conventions are accounted for. This finding is significant because it suggests that synthetic data can be a useful replacement for real data. Furthermore, we were able to achieve this result using only 22k images, which is similar in size to the real dataset but required significantly less effort in manual collection and no effort to annotate.

We also demonstrate that the distribution gap makes up a larger segment of the performance gap.
At the start of our study, we discovered a significant difference in performance between a model trained on our initial dataset and one trained on real data. Specifically, there was a gap of 10.5 percentage points between the F$_1$ score on real data for the two models, with the model trained on our initial dataset achieving a score of 80.7\% and the model trained on real data achieving a score of 91.2\%. We identified the distribution gap as the primary contributor to this performance gap. Adding more content to align the distributions reduced 53\% of this gap, by increasing the F$_1$ score 5.6 percentage points. The label gap likely accounted for another 36\% of this gap, or 3.8 percentage points in F$_1$ score. It's worth noting that the label adaptation model could potentially correct some segmentation errors in addition to addressing the label gap, so this number might be slightly lower. The remaining 10\% of the gap, or 1.1 percentage points in F$_1$ score, can be attributed to both the photorealism gap and additional differences in the distribution that we assume were not perfectly addressed.

We observe that increasing the size of the fine-tuning dataset leads to an increase in the score until we reach the point where fine-tuning with the full real dataset and training from scratch both yield equivalent results. Our results also demonstrate that regardless of the amount of real data used, fine-tuning a model trained on synthetic data consistently yields better results than training solely on the equivalent amount of real data. In addition, we show that fine-tuning with a small amount of real data---as little as 1\% of the total dataset or 200 images---can be helpful in improving the performance of a synthetically trained model.

To understand the performance gaps, we tested both label adaptation and fine-tuning. However, in a production-oriented system with a fixed set of real, annotated images, we recommend performing fine-tuning over label adaptation. Fine-tuning is simpler and more effective in handling all three gaps.

\section{Future Work}
\label{section:future_work}
Our study has demonstrated that aligning the distribution and labels between synthetic and real data can lead to competitive results comparable to training purely on real data. However, we believe that the remaining distribution and photorealism gaps can be further closed by adding more variance to our dataset. It would also be interesting to explore the impact of training with a significantly larger dataset.
In addition, future work could focus on developing a better metric to precisely measure the size of the different gaps - the distribution gap, the photorealism gap, and the label gap. This would enable a more precise evaluation of the effectiveness of different techniques for closing these gaps.
Moreover, while we manually aligned the distribution of the synthetic dataset with the real data, there is still scope for further automation of this process through data-centric iteration to close the distribution gap.

\bibliographystyle{unsrtnat}
\bibliography{references}

\appendix

\section{Augmentations}
\label{appendix:augmentations}
We implemented the augmentations using the Albumentations library \cite{albumentations}.
\begin{itemize}
    \item \textbf{CelebAStyleAlignment} Align the face so that the eyes are in the center of the image. This is the same algorithm that is used to align the images in the CelebA dataset to create the CelebAMask dataset. (We apply it with probability 0.3)
    \item \textbf{HorizontalFlip} Randomly flip the image horizontally (with probability 0.1)
    \item \textbf{ShiftScaleRotate} Randomly shift the image, scale it, or slightly rotate it (with probability 0.1)
    \item \textbf{Perspective} Apply a perspective transform (with probability 0.1)
    \item \textbf{Blur} Randomly blur the image (with probability 0.1)
    \item \textbf{RandomBrightnessContrast} Randomly change brightness and contrast of the image (with probability 0.05)
    \item \textbf{GaussianNoise} Randomly add gaussian noise to the image (with probability 0.1)
    \item \textbf{ToGray} Randomly add gaussian noise to the image (with probability 0.1)
    \item \textbf{RandomFog} Randomly simulate fog on the image (with probability 0.1)
    \item \textbf{RandomShadow} Randomly simulate shadow on the image (with probability 0.1)
    \item \textbf{RandomRain} Randomly simulate rain on the image (with probability 0.1)
    \item \textbf{Spatter} Randomly simulate spatter marks as from mud or rain on the image (with probability 0.1)
    \item \textbf{HueSaturationTransform} Randomly shift the hue and saturation of the image (with probability 0.1)
    \item \textbf{ImageCompression} Randomly simulate JPEG compression on the image (with probability 0.5). Datagen's images are generated as lossless png images whereas the images from CelebAMask are lossy jpeg images.
\end{itemize}

\section{Result Comparison Table}
\label{appendix:result_table}
We show the same data as in Figure \ref{fig:results_comparison}, but in table form. Note that the label adaptation experiments were not completed for 8,000 and 16,000 images. For reference, training on synthetic data without label adaptation achieved a score of 86.3\%

\centering
\begin{tabular}{llll}
\toprule
\# Images & Synthetic + Fine-tune & CelebA & Synthetic + Label Adaptation \\
\midrule
200 & 89.4 & 85.1 & 87.3 \\
400 & 89.7 & 84.5 & 87.8 \\
800 & 90.1 & 87.5 & 88.5 \\
1000 & 90.2 & 88.0 & 88.8 \\
2000 & 90.5 & 88.9 & 89.0 \\
4000 & 90.8 & 89.8 & 89.3 \\
8000 & 91.0 & 90.5 & NaN \\
16000 & 91.1 & 90.9 & NaN \\
24000 & 91.3 & 91.2 & 90.0 \\
\bottomrule
\end{tabular}

\end{document}